\documentclass[runningheads]{llncs}
\usepackage{graphicx}
\usepackage{amsmath,amssymb} 
\usepackage{color}
\usepackage{amsfonts}
\usepackage{tabularx,multirow}
\usepackage{epstopdf}

\begin{document}

\title{Automatic Graphics Program Generation using Attention-Based Hierarchical Decoder\thanks{This work was supported by the National Key R\&D Program of China (No.2016YFB1001001) and the National Natural Science Foundation of China (No.91648121, No.61573280).}} 
\titlerunning{Auto GUI Code Generation} 


\author{Zhihao Zhu \and
Zhan Xue \and
Zejian Yuan}
%

\authorrunning{Z. Zhu et al.} 


\institute{Institute of Artificial Intelligence and Robotics\\
 School of Electronic and Information Engineering, Xi'an Jiaotong University\\
 Xi'an, China\\
\email{zzh123@stu.xjtu.edu.cn}~~~~\email{xx674967@stu.xjtu.edu.cn}~~~~  \email{yuan.ze.jian@xjtu.edu.cn}\\
}

\maketitle

\begin{abstract}
Recent progress on deep learning has made it possible to automatically transform the screenshot of Graphic User Interface (GUI) into code by using the encoder-decoder framework. While the commonly adopted image encoder (e.g., CNN network), might be capable of extracting image features to the desired level, interpreting these abstract image features into hundreds of tokens of code puts a particular challenge on the decoding power of the RNN-based code generator. Considering the code used for describing GUI is usually hierarchically structured, we propose a new attention-based hierarchical code generation model, which can describe GUI images in a finer level of details, while also being able to generate hierarchically structured code in consistency with the hierarchical layout of the graphic elements in the GUI. Our model follows the encoder-decoder framework, all the components of which can be trained jointly in an end-to-end manner. The experimental results show that our method outperforms other current state-of-the-art methods on both a publicly available GUI-code dataset as well as a dataset established by our own.
\end{abstract}

\section{Introduction}

Using machine learning technologies to generate codes of Graphic User Interface automatically is a relatively new field of research. Generally, implementing the GUI is often a time-consuming and tedious work for the front-end developers, which disputes them from devoting more time to developing the real functionalities and logics of the software. So developing systems to transform the GUI mockup into programming code automatically shows a very promising application potential.

Recent examples of using machine learning technologies to automatically generate programs in a human-readable format are \cite{C3,C8}, who use gradient descent to induce source code from input-output examples via differentiable interpreters. However, their performance has been proven by \cite{C4} to be
\begin{figure}[htb]
\centerline{\includegraphics[width=12.5cm]{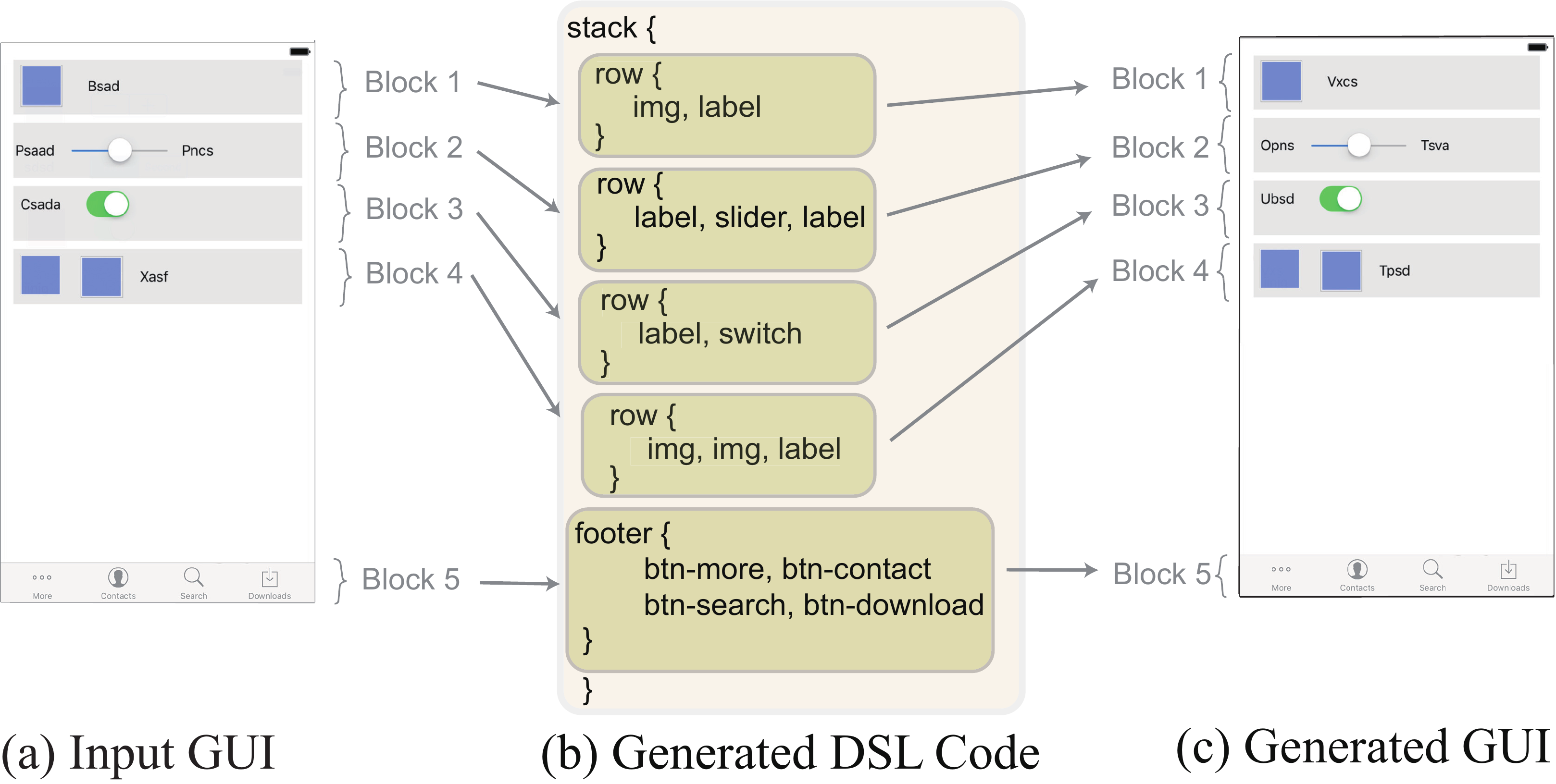}}
\caption{An Example of the code generated by our model given an input GUI screenshot. (a) gives an example of how a GUI is divided into blocks. (b) shows the generated DSL code. (c) is a rendered GUI based on the generated DSL code. }
\label{f1}
\end{figure}
inferior to discrete search-based techniques that are widely used by the programming language community. Another work is the DeepCoder\cite{C1}, a system attempting to generate computer programs by leveraging statistical predictions to augment traditional search techniques. Its ability to model complex programming languages is however limited by its reliance on the \emph{Domain Specific Languages (DSL)}, which are designed only for a specialized domain.


In the task of generating codes from visual inputs, there has been only a few numbers of works \cite{C50,C2}, among which, pix2code \cite{C2} is the most similar to ours. \cite{C2} employs a feedback mechanism, where decoding process can be proceeded iteratively with two different levels of LSTMs: an ``encoding'' LSTM for encoding the already generated code sequence to reduce the ``decoding'' LSTM's burden for learning long sequence relationships. The ``decoding'' LSTM is for the code sequence generation, and it then feeds the decoded token sequence back into the ``encoding'' LSTM to form a feedback loop. By employing the feedback mechanism, \cite{C2} is able to generate much longer word/token sequence than other single-pass based methods. However, their method needs to pre-fix the maximum sequence length that the ``encoding'' LSTM can generate. In other words, the range of the code length should be pre-specified, which reduces their method's extensibility and generalization ability. Another apparent defect of their methods is that they didn't take into consideration the hierarchical structure of GUI and its code, which limits their performance in generating accurate graphics programs.

To tackle the above problems, we propose a new method for automatic graphics program generation. It not only can well solve the long-term dependency problem, but can also capture the hierarchical structure of the code by explicitly representing the code generation process in a hierarchical approach. Our method reasons about the code sequence using a hierarchical decoder, generating graphic programs block by block. Figure \ref{f1} shows an example of GUI and its corresponding code, as well as the approach for dividing the GUI into blocks. The detailed DSL code generation process is as follows: first, a first-stage LSTM is used to decode image's visual contents in the block level, with its every hidden state containing general context information of that block. Then, we feed the first-stage LSTM's hidden state as well as image's convolutional features into an attention model to select the most important parts of the convolutional features, which are then input into the second-stage LSTM as context information to generate code tokens for the corresponding block.

The experimental results on a benchmark dataset provided by \cite{C2} well demonstrate the effectiveness of our method: we perform better than current state-of-the-art methods across all three sub-datasets: iOS, Android, and Web. Besides, to further illustrate our proposed model's advantage in dealing with more complex GUIs, we establish a new dataset containing GUI screenshots that have more graphic elements and more diversity in graphic elements' style and spatial layout. On this dataset, our method outperforms the compared methods even by a larger margin.

\section{Related Work}
Automatically generating code for the GUI screenshot is very similar to the task of image captioning \cite{C6,C5,C9,C10,C30,C102,C101}, both of which need first to understand image's visual contents, and then interpret them into the language form. Our proposed method for generating GUI code follows the encoder-decoder framework and incorporates the attention mechanism. So we mainly introduce the related work about automatic GUI code generation with them.

Most recent neural network-based methods for generating image descriptions follow the encoder-decoder framework, which first uses a deep CNN to encode an image into an abstract representation, and then uses a RNN to decode that representation into a semantically-meaningful sentence which can describe the image in details. Chen et al. \cite{C21} learned a bi-directional mapping between images and their sentence-based descriptions using RNN. Mao et al. \cite{C12} propose a Multimodal Recurrent Neural Network (MRNN) that uses an RNN to learn the text embedding, and a CNN to learn the image representation.  \cite{C17} and \cite{C18} show that high-level semantic attributes can boost the image caption generation performance. Inspired by the successful application of attention mechanism in machine language translation \cite{C14}, spatial attention has also been widely adopted in the task of image captioning \cite{C15,C16,C19}. However, due to the limited capability of a single LSTM to learn the long sequence dependency, the above methods are not able to generate very long image descriptions. It's also a bottleneck of these methods for interpreting more complex GUIs into codes, which usually contain hundreds of thousands of tokens.

\begin{figure}[htb]
\centering
\includegraphics[width=12.5cm]{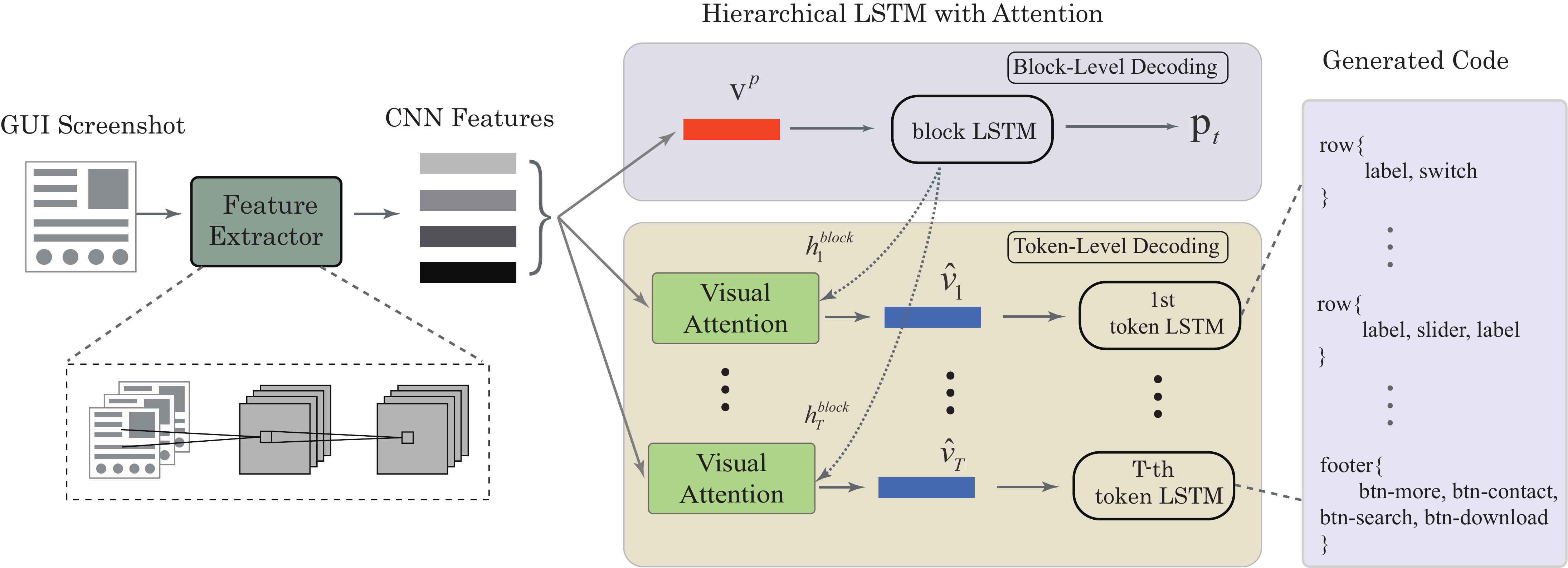}
\caption{Overview of our proposed model for automatic graphics program generation. The input GUI screenshot is first fed into a CNN to obtain high-level visual features. All the visual features are projected to $\mathbb R^{D}$, pooled to give a compact image representation, and is then fed into block LSTM as input. Block LSTM determines the number of blocks to generate based on $p_t$, and generates the guiding vector $\boldsymbol{h}_t^{block}$, which is fed into the attention model for selecting CNN features to feed into $t$-th token LSTM to generate codes for the $t$-th block.   }
\label{2}
\end{figure}
To tackle the above issue, Jonathan et al. \cite{C51} employ the hierarchically structured LSTMs, and their model is able to generate paragraph-level image descriptions. Basically, in their work, two levels of LSTM-based language decoders are used: a first-stage LSTM captures the general information of the image and stores context information of each sentence in its hidden states. Then, a second-stage LSTM is used to decode the first-stage LSTM's hidden states into different sentence in a paragraph.
Our method for automatic GUI code generation also employs a similar hierarchical approach. However, in our method, first-stage LSTM's hidden states are only used in the attention network to help select the CNN features, which are then fed into the second-stage LSTM for code generation. Our experimental results show that the first-stage LSTM's hidden states, which only contain a coarse level of visual information, are insufficient for guiding the second-stage LSTM to generate accurate token sequences. We do obtain better results by feeding the more information-carrying raw CNN features into the second-stage LSTM.

The contributions of our work are as follows: (1) We are the first to propose using a hierarchical approach for generating programs from the graphics user interface. (2) We propose a new method to integrate attention mechanism with hierarchical LSTM, which outperforms current state-of-the-art methods that also use hierarchical LSTM. (3) We introduce a new dataset named PixCo-e \emph{(details in Section 4.1)}, which contains GUI-code examples from three different platforms: iOS, Android, and Web.

\section{Proposed Method}
\subsection{Overview}

Our method takes a GUI screenshot as input, generating a body of programming languages to describe it, and is designed to take advantage of the hierarchical structure of both the GUI and the graphics programs. Figure \ref{2} provides an overview of our attention-based hierarchical code generation model. We first use the intermediate filter responses from a Convolutional Neural Network (CNN) to build a high-level abstract visual representation of the image, denoted by $\boldsymbol{\nu}$, which are then taken as input by a hierarchical LSTM composed of two levels: a block LSTM and a token LSTM. The block LSTM receives the image's visual features as input, and decides how many code blocks to generate in the resulting program. The hidden state $\boldsymbol{h}_{t}^{block}$ of the block LSTM at each time step $t$ is then fed into an attention model as a guiding vector to select the most important sub-regions in the visual feature map. Given the selected visual features as context, a token LSTM generates the code for the corresponding block.

\subsection{Vision Encoder: Convolutional Features}
We first follow \cite{C2} to design a lightweight \emph{DSL} to describe GUIs. Generally, our method generates a computer program as a long
sequence of 1-of-\emph{K} encoded tokens:
\begin{equation*}
  q = \{\mathbf{q}_1,\mathbf{q}_2,...\mathbf{q}_C\},~~~{\mathbf{q}_i}\in{\mathbb R}^K,
\end{equation*}
where $C$ is the total number of tokens in the program, and \emph{K} is the size of the token vocabulary.

We use a convolutional neural network as the image encoder to extract a set of visual feature vectors $\nu = [ {\mathbf{v}_1},{\mathbf{v}_2},...,{\mathbf{v}_L}]$, and each $\mathbf{v}_i$ is a $D$-dimensional representation corresponding to a certain part of the image.
To obtain a correspondence between the feature vector and each sub-region in the 2-D image, we extract features from a lower convolutional layer instead of using the final fully connected layer.
\\
\\
\textbf{Region Pooling:} We want to aggregate all the visual features $ \nu =[\mathbf{v}_1, \mathbf{v}_2,..., \mathbf{v}_L]$ into a single pooled vector $\mathbf{v}^p\in{\mathbb R}^D$ that compactly describes the content of the image, which later will be fed into the block LSTM as context. The pooled vector $\mathbf{v}^p$ is computed by taking an element-wise maximum on each channel: $v^{p,c} = {max}\{v_1^c, v_2^c,..., v_L^c\}$, where $v^{p,c}$ and $v_i^c$ represent the $c$-th channel of $\mathbf{v}^p$ and $\mathbf{v}_i$, respectively.

\subsection{Attention-Based Hierarchical Decoder}
The extracted visual features are fed into a hierarchical visual decoder model composed of two modules: a block LSTM and a token LSTM. The block LSTM is responsible for determining the total number of blocks that the entire program will have, and for generating a \emph{H}-dimension guiding vector for each of these blocks. Given the guiding vector, the token LSTM then uses the attention mechanism to select the most important visual features as the context information at each time step to generate code tokens of that block.

\subsubsection{Block LSTM}
The block LSTM is a single-layer LSTM with a hidden size of \emph{H} = 512. Its initial hidden state and cell state are both set to zero. The block LSTM receives the pooled feature vector $\mathbf{v}^p$ as the initial input, and in turn produces a sequence of hidden states $\boldsymbol{H}^{block}$ = [$\boldsymbol{h}_{1}^{block}$, $\boldsymbol{h}_{2}^{block}$,$...$], each corresponds to a single block in the program.
 The overall working flow of the block LSTM is governed by the following equations:
\begin{equation}\label{1}
{\boldsymbol{x}_0} = {W_{x,v}} \cdot \mathbf{v}^p,
\vspace{1em}
\end{equation}
\begin{equation}\label{2}
{\boldsymbol{x}_t} = {W_{x,o}} \cdot {\boldsymbol{o}_{t - 1}},
\vspace{1em}
\end{equation}
\begin{equation}\label{3}
{\boldsymbol{h}_t^{block}} = LSTM({\boldsymbol{h}_{t - 1}^{block}},{\boldsymbol{x}_t}),
\vspace{1em}
\end{equation}
\begin{equation}\label{4}
{\boldsymbol{o}_t} = \sigma (W_{o,h} \cdot {\boldsymbol{h}_t^{block}}),
\end{equation}
where $W$s denote weights. $\sigma(\cdot)$ stands for sigmoid function, and ${\boldsymbol{o}_t}$ is the output of block LSTM at each time step. For clearness, we do not explicitly represent the bias term in our paper.

Each hidden state $\boldsymbol{h}_{t}^{block}$ is used in two ways: First, a linear projection from $\boldsymbol{h}_{t}^{block}$ with a logistic classifier produces a distribution $p_t$ over the two states \{CONTINUE = 0, STOP = 1\}, which determine whether the $t$-th block is the last block in the program. Second, the hidden state $\boldsymbol{h}_{t}^{block}$ is fed into an attention network to select specific CNN features, which are input into the following token LSTM as context information. Each element $\boldsymbol{h}_{i}^{block}$ in $\boldsymbol{h}^{block}$ contains information about the input visual features with a strong focus on parts surrounding each specific block.

\subsubsection{Token LSTM}
Once the block LSTM's hidden states are obtained, we use the following approach to select the CNN features to feed into the token LSTM:
We first reshape the convolutional visual features $\nu = [ {\mathbf{v}_1},{\mathbf{v}_2},...,{\mathbf{v}_L}]$ by flattening its width $W$ and height $H$, where $L = W \cdot H$. Then, we use a multi-layer perceptron with a softmax output to generate the attention distribution $\boldsymbol{\alpha} = \{\alpha_1, \alpha_2,...,\alpha_L\}$ over the image regions. Mathematically, our attention model can be represented as:
\begin{equation}\label{7}
  \boldsymbol{e}_{t} = {f_{MLP}}(({W_{e,\boldsymbol{v}}}\boldsymbol{v}) + ({W_{e,h}}{h_{t}^{block}}))
  \vspace{0.8em}
\end{equation}
\begin{equation}\label{8}
  \boldsymbol{\alpha}_{t}  = softmax({W_s}\boldsymbol{e}_t)
\end{equation}
\begin{equation}\label{9}
 \widehat {\mathbf{v}_{t}} = \sum\limits_{i = 1}^L {{\alpha_{i,t}}{\mathbf{v}_i}}
\end{equation}
Where $f_{MLP}(\cdot)$ represents a multi-layer perceptron. We follow the ``soft" approach as illustrated in Equation (7) to gather all the visual features using the weighted sum, and $\widehat {\mathbf{v}_{t}}$ stands for the selected visual features that will be fed into the $t$-th token LSTM as context information.

The token LSTM is a two-layered LSTM with hidden size of $H = 512$. Notice that there are multiple token LSTMs, and the total number of token LSTM equals the total number of blocks. Each token LSTM receives the selected visual features $\widehat {\mathbf{v}}$ as its initial input, and is responsible for generating code for a block.
Token LSTM's subsequent inputs are the embedding form of the output token generated in the previous time step. The hidden state $\boldsymbol{h}_{t}^{{token}}$ of the second LSTM layer is used to predict a distribution over the tokens in the vocabulary, and a special END token signals the end of a block.

After each token LSTM has generated the codes of their corresponding blocks, these blocks of code are concatenated to form the complete GUI program.

\subsection{Training}
The use of hierarchical LSTM makes it possible to train the entire model in an end-to-end manner. Training data consists of $(x, y)$ pairs, where $x$ represents the input GUI screenshot and $y$ is the code for
that GUI. Different from \cite{C2}, who used a fix-sized sliding window to obtain slices of code to feed into their ``encoding'' LSTM at different training iteration, our method only needs to feed the entire body of code into the model once.

Because we use the same \emph{DSL} as the intermediate programming language to describe GUIs across all three platforms, we follow a consistent rule to divide the codes in the training dataset into blocks: we first ignore the ``stack'' token at the beginning of each program body, as well as its corresponding open brace ``\{" and ending brace ``\}". Besides of these two special braces, each one of the other open braces signals the beginning of a block, and the corresponding ending brace signals the end of that block. We then manually insert a BLOCK-END token at the end of each block. After the division, we assume that the whole program contains $S$ blocks, and the $i$-th block has $N_i$ tokens, and $y_{ij}$ is the $j$-th token of the $i$-th block.

We then unroll the block LSTM for $S$ time steps, providing a distribution $p_i$ over the \{CONTINUE = 0, STOP = 1\} states for each block. Each block LSTM's hidden state is used for selecting the visual features, which are later fed into $S$ copies of token LSTM to produce distribution $q_{ij}$ over each token. Our training loss for the example pair $(x, y)$ is a sum of two cross entropy functions: a block loss $l_{block}$ on the stopping distribution $p_i$, and a token loss $l_{token}$ on the token distribution $q_{ij}$:
\begin{equation}\label{10}
 L = \sum\limits_{i = 1}^S{l_{block}(p_i,g_i)}+\sum\limits_{i = 1}^S \sum\limits_{j = 1}^{N_i}{l_{token}(q_{ij},y_{ij})}
\end{equation}
where $g_i$ is the groundtruth distribution of the \{CONTINUE = 0, STOP = 1\} states for $i$-th block, and $y_{ij}$ is the groundtruth distribution over $j$-th token in $i$-th block.

\section{Experiment}

\subsection{Setup}

\textbf{Data:}~~~We implement the proposed automatic graphic program generation model on two datasets, both of which are composed of GUI screenshots and the corresponding source codes.

\begin{figure}[htb]
\centerline{\includegraphics[width=12.5cm]{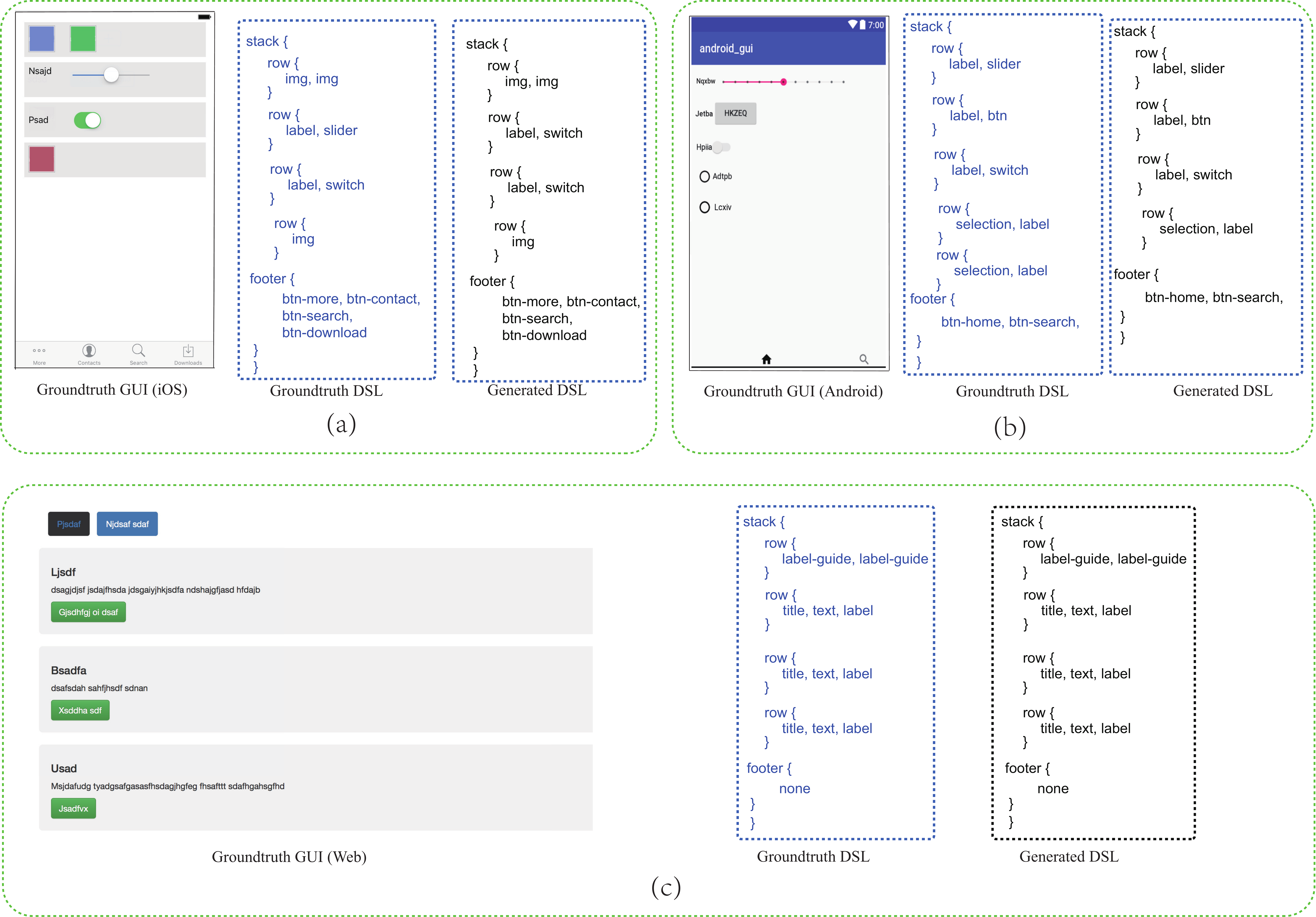}}
\caption{Qualitative results of our model on the PixCo dataset. DSL code generated by our method(black), groundtruth DSL code(blue), GUI screenshots are shown. (a), (b), (c) represent examples from iOS, Android and Web platforms respectively.}
\label{f3}
\end{figure}

\textbf{(1)} The first one is a public dataset \footnote{https://github.com/tonybeltramelli/pix2code/tree/master/datasets} provided by \cite{C2}. For readability, we denote this dataset as PixCo dataset. It consists of three sub-datasets, corresponding to the GUI-code pairs in three different platforms: iOS, Android, and Web. Each sub-dataset has 1500 GUI-code pairs for training and 250 GUI-code pairs for testing.

\textbf{(2)} The second one is our own dataset: PixCo-e dataset. The reason for establishing this new dataset is that the GUI examples in the PixCo dataset are relatively too simple, while the real-world GUIs contain much more graphic elements, and have more complex visual layout. The GUIs in our PixCo-e dataset keep all the basic graphic elements used in the PixCo dataset, such as labels, sliders, switches, but the difference is that each GUI in our PixCO-e dataset contains much more graphic elements and the visual layout is more complex\emph{(examples can be seen in figure \ref{web} and figure \ref{ios})}. Similar to PixCo, PixCo-e is also composed of three sub-datasets of GUI-code pairs corresponding to iOS, Android, and Web platforms. Each subset is comprised of 3000 GUI-code pairs for training and 500 GUI-code pairs for testing.

Note that we follow the same rule illustrated in \emph{section 3.4} to divide the codes into blocks for both datasets.\emph{(The code and dataset are published in \footnote{https://github.com/ZhihaoZhu/Auto-GUI-Code-Generation})}
\begin{figure}[htb]
\centerline{\includegraphics[width=12.5cm]{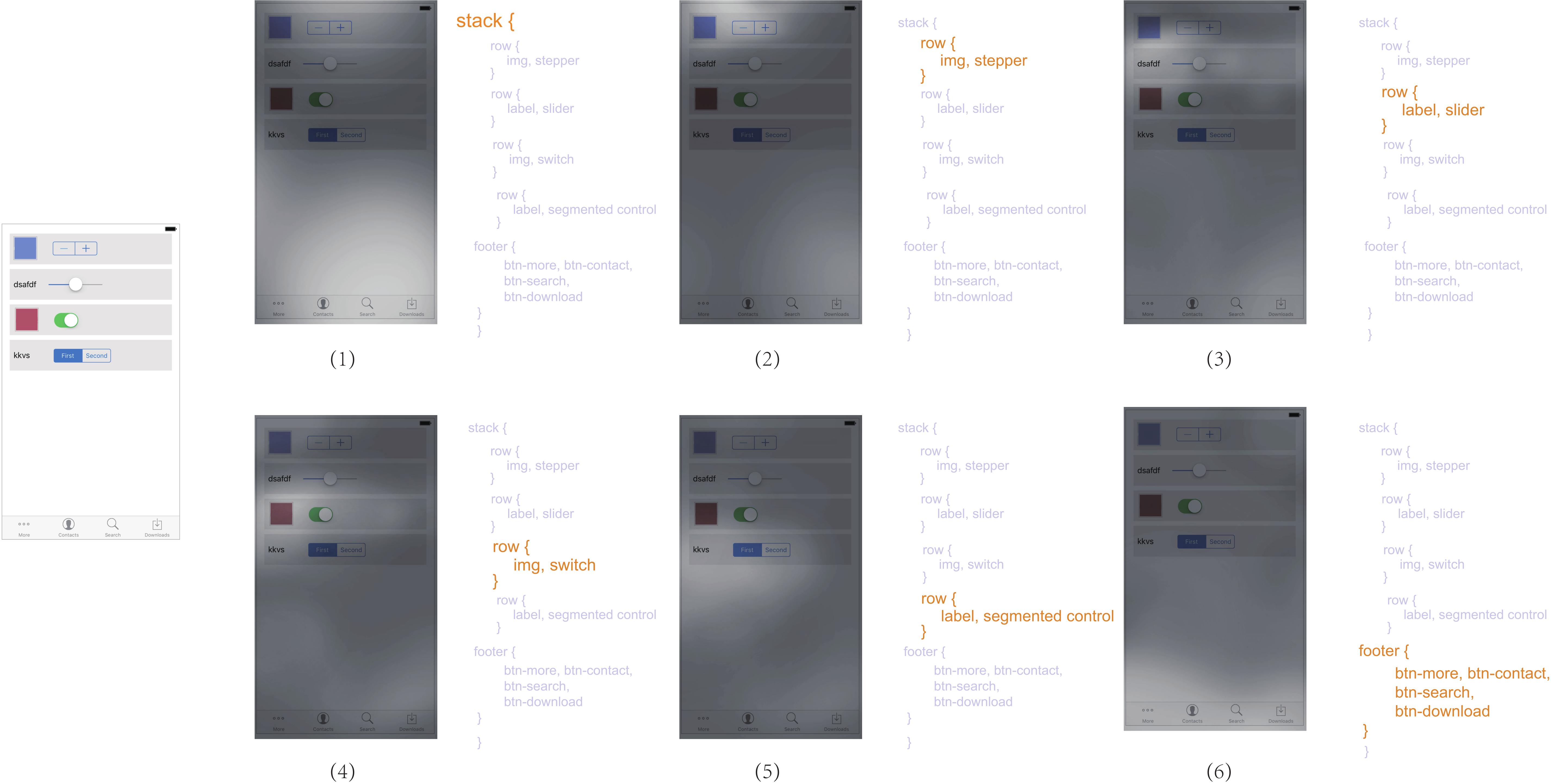}}
\caption{Attention over time. As the model generates each block, its attention changes to reflect the relevant parts of the image. Brighter areas indicate where the attentions are paid, while darker areas are less attended. The orange noted codes are the blocks generated at the current step.}
\vspace{-1.5em}
\label{f15}
\end{figure}

\textbf{Implementation Details:}~~~We first re-size the input images to 256 $\times$ 256 pixels and then the pixel values are normalized.
For the encoding part, for a fair comparison, we employ the same CNN-based encoder architecture adopted by \cite{C2}, which consists of three convolutional layers, with widths of 32, 64 and 128, respectively, and two fully-connected layers. For the decoding part, the dimensions of both block LSTM's and token LSTM's inputs are set to 512, and \emph{sigmoid} is used as the nonlinear activation function. Moreover, as the size of our token vocabulary is small, we directly use the token-level language modeling with a discrete input of one-hot encoded vector.

In the training procedure, we use Adam algorithm\cite{C7} for model updating with a mini-batch size of 128. We set the model's learning rate to 0.001 and the dropout rate to 0.5. Using a single NVIDIA TITAN X GPU, the whole training process takes about three hours on the PixCo dataset and six hours on our PixCo-e dataset.
\\
\\
\textbf{Baseline Methods:}~~~To demonstrate the effectiveness of our method, we also present the results generated by two baseline methods. Notice that two baseline methods are trained and tested on the same datasets, and all the hyper-parameters and CNN architectures remain the same for the two baseline models.

\emph{(Baseline-1:)} The first baseline method follows a simple encoder-decoder architecture, where a CNN is used to encode the pixel information into the high-level visual features, and a single-layer LSTM is used to decode these features into the token sequence.

\begin{table*}
\centering
\begin{tabular}{|c|c|c|c|c|c|c|}
\hline
\multicolumn{1}{|c|}{\multirow{3}{*}{Dataset Type}} & \multicolumn{6}{c|}{Accuracy of Block Partitioning (\%)}\\
\cline{2-7}
\multicolumn{1}{|c|}{} & \multicolumn{3}{c|}{PixCo} &\multicolumn{3}{c|}{PixCo-e}\\
\cline{2-7}
\multicolumn{1}{|c|}{} & \textbf{~~iOS~~} & \textbf{Android} & \textbf{~~Web~~} & \textbf{~~iOS~~} & \textbf{Android} & \textbf{~~Web~~}\\
\hline
\hline
Baseline-2 &  90.2 & 91.4 & 93.3 & 84.4 & 85.1 & 90.0 \\
\hline
Ours & 90.7 & 91.9 & 94.0 & 85.1 & 85.9 & 91.5 \\
\hline

\end{tabular}
\vspace{1em}
\caption{Comparison of the performance in block partitioning accuracies on PixCo and PixCo-e datasets by baseline-2 method and our method.}
\vspace{-1em}
\label{5}
\end{table*}

\emph{(Baseline-2:)} The second baseline method follows the hierarchical LSTM-based architecture. It also first encodes the image's pixel information into high-level visual features. Then, the visual features are pooled to obtain a more compact representation, which is then fed into a two-stage hierarchical LSTM.
Compared to ours, the only difference of the hierarchical LSTM used in baseline-2 is that the first-stage LSTM's hidden states at each time step are directly fed into the second-stage LSTM as context, without passing through any other modules.


\subsection{Effect of Block LSTM}
As both of our method and the baseline-2 method generate programs in a block-by-block way, we first want to test two methods' accuracy in partitioning programs into the right number of blocks. We calculate the block partitioning accuracy $A_{bp}$ by following equations:
\begin{equation}\label{10}
 A_{bp} = \frac{1}{G}\sum\limits_{i = 1}^G{f(c_p^i-c_{gt}^i)},
\end{equation}
\begin{equation}\label{10}
  f(x)=\begin{cases}
    1,\quad |x| = 0 \\
    0,\quad other
\end{cases} ,
\end{equation}
where $G$ represents the number of examples in a dataset. $c_p^i$ is the generated number of blocks, and $c_{gt}^i$ is the ground-truth number of blocks.

Table. \ref{5} illustrates the block partitioning accuracies achieved by the baseline-2 method as well as our method. We note that two methods can both partition the program into blocks in a satisfying manner. And the performance of two methods are very close, which can be explained as follows: two models' ability in correctly partitioning the programs is mainly trained via minimizing the block loss $l_{block}$ as in Equation (8). During training, this loss is directly passed to the first-stage LSTM without going through models' other modules (e.g., Attention network, second-stage LSTM...), where lies two models' main structural differences. Thus, by trained on the same dataset, the first-stage LSTMs in two models are equipped with very similar capabilities in correctly partitioning programs into blocks.

\subsection{Quantitative Evaluation Results}
\begin{table*}
\centering
\begin{tabular}{|c|c|c|c|c|c|}
\hline
\multicolumn{2}{|c|}{\multirow{2}{*}{Dataset Type}} & \multicolumn{4}{c|}{Error (\%)}\\
\cline{3-6}
\multicolumn{2}{|c|}{} & pix2code \cite{C2} & baseline-1 & baseline-2 & Ours\\
\hline
\hline
\multirow{3}*{PixCo} & iOS & 22.73 & 37.94 & 20.90 & \textbf{19.00}  \\
\cline{2-6}
& Android& 22.34 & 36.83 & 21.72 & \textbf{18.65} \\
\cline{2-6}
& Web& 12.14 & 29.77 & 11.92 & \textbf{11.50} \\
\hline
\hline

\multirow{3}*{PixCo-e} & iOS & 36.4 & 53.21 & 29.43 & \textbf{26.79} \\
\cline{2-6}
& Android& 36.94 & 54.75 & 29.11 & \textbf{26.10} \\
\cline{2-6}
& Web& 29.47 & 43.60 & 22.70 & \textbf{18.31}  \\
\hline

\end{tabular}
\vspace{1em}
\caption{Performance of our proposed method on the test dataset, comparing with the baseline-1, baseline-2 and pix2code\cite{C2} methods. All four methods use the greedy search strategy.}
\vspace{-1.5em}
\label{3}
\end{table*}

\begin{table*}
\centering
\begin{tabular}{|c|c|c|c|c|}
\hline
\multicolumn{2}{|c|}{\multirow{3}{*}{Dataset Type}} & \multicolumn{3}{c|}{Error (\%)}\\
\cline{3-5}
\multicolumn{2}{|c|}{} & \multicolumn{3}{c|}{Ours}\\
\cline{3-5}
\multicolumn{2}{|c|}{} & \textbf{greedy} & \textbf{beam-3} & \textbf{beam-5}\\
\hline
\hline
\multirow{3}*{PixCo} & iOS & 19.00 & \textbf{17.90} & 17.43\\
\cline{2-5}
& Android& 18.65 & 18.20 & \textbf{16.31}\\
\cline{2-5}
& Web& 11.50 & 11.23 & \textbf{10.88}\\
\hline
\hline
\multirow{3}*{PixCo-e} & iOS  & 26.79 & \textbf{23.40} & 23.62\\
\cline{2-5}
& Android& 26.10 & 27.20 & \textbf{26.78}\\
\cline{2-5}
& Web& 18.31 & 17.25 & \textbf{16.11}\\
\hline

\end{tabular}
\vspace{1em}
\caption{Comparison of our method's performance by using different search strategies: ``greedy'' represents greedy-based method, ``beam-3'' and ``beam-5'' stand for beam search strategy using beam size of 3 and 5, respectively.}
\vspace{-2.5em}
\label{4}

\end{table*}

Table. \ref{3} compares our method to the pix2code\cite{C2} and two baseline methods on the PixCo dataset as well as our PixCo-e dataset. The quality of the generated code is evaluated by computing the classification error for each sampled token. While the length of the generated token sequence might differ from the ground truth, we still count those unmatched tokens as errors.

When comparing our method with pix2code \cite{C2}, we first note that our method outperforms theirs across all three sub-datasets on the PixCo dataset. It's worth noting that on the web sub-dataset, the performance difference between the two methods is very minor. Actually, the average \emph{HTML/CSS} code length used to implement web GUI is much shorter (37 tokens/GUI) than the \emph{storyboard} code for iOS (51 tokens/GUI) and \emph{XML} code for Android (53 tokens/GUI). The better performance in the iOS/Android platforms who have longer code length demonstrates our model's advantage in interpreting more complex GUI and handling longer sequence dependency. This can be further illustrated by our model's even lower error rates across all three platforms in the PixCo-e dataset. Note that compared to the PixCo dataset, the average code length in PixCo-e dataset is much longer: iOS (109 tokens/GUI), Android (112 tokens/GUI) and web (86 tokens/GUI).

\begin{figure}[htb]
\centerline{\includegraphics[width=11.5cm]{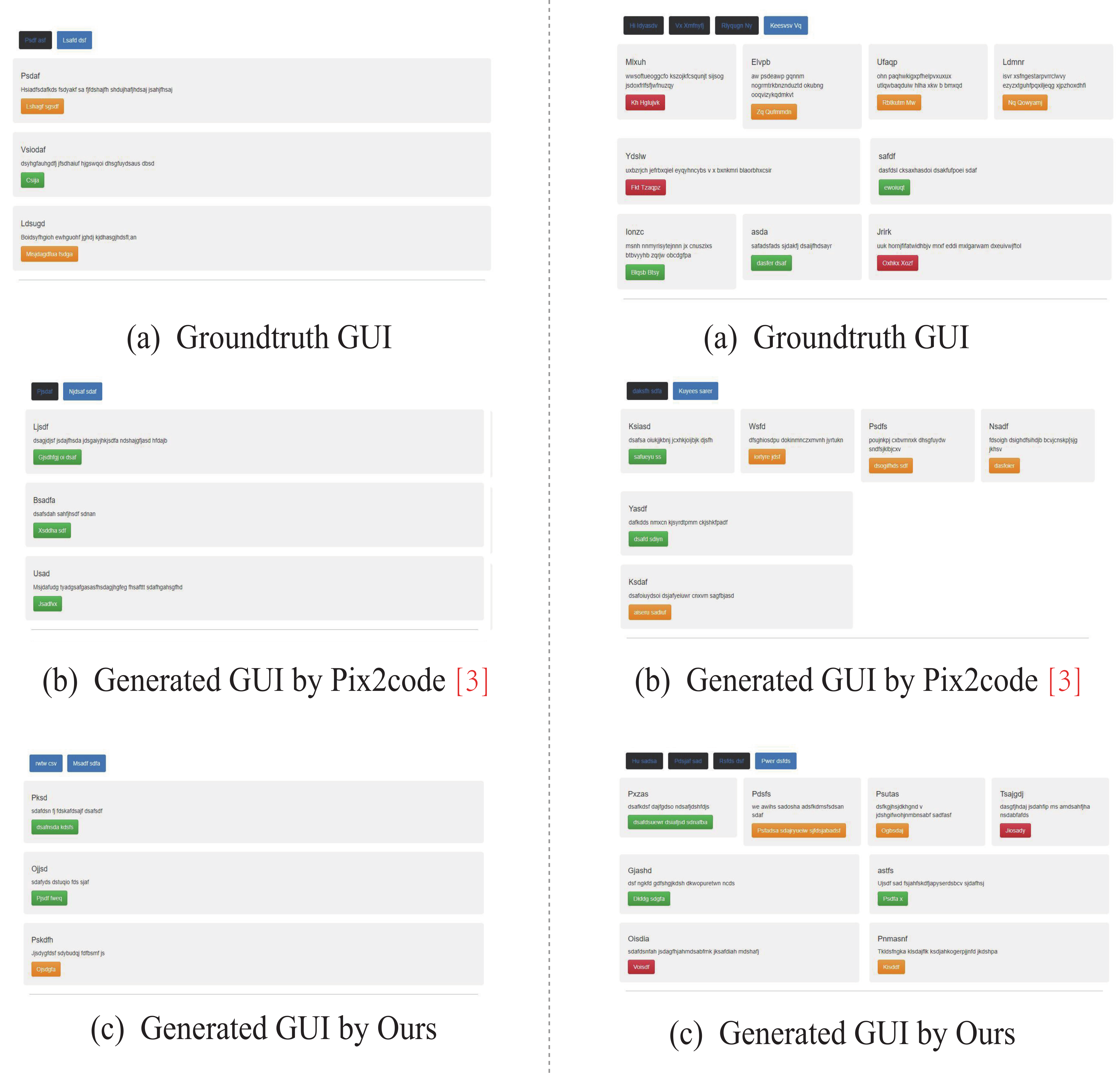}}
\caption{Experiment samples from the web-based GUI dataset. The examples on the left column are form the PixCo dataset and the examples on the right column are from the PixCo-e dataset. }
\label{web}
\end{figure}

When comparing our method to the baseline-2 method, which also employs a hierarchical architecture, we note that our method still performs better across all the platforms in both two datasets. This well demonstrates that when using the same CNN, adding attention mechanism can boost the hierarchical model's performance in generating accurate token sequences.

Table. \ref{4} compares the performance of our method when using different search strategies during sampling. We note that better performance is achieved by using the beam search strategy.

\subsection{Qualitative Evaluation Results}

Figure \ref{f3} shows examples of the GUI screenshots, ground-truth DSL codes as well as the DSL codes generated by our model. What's more, we also show in Figure \ref{f15} the intermediate dynamic attention process during the code generation. As we can see, the model is able to learn proper alignments between the code blocks and their corresponding spatial areas in the GUI screenshot.

\begin{figure*}[htb]
\centerline{\includegraphics[width=11cm]{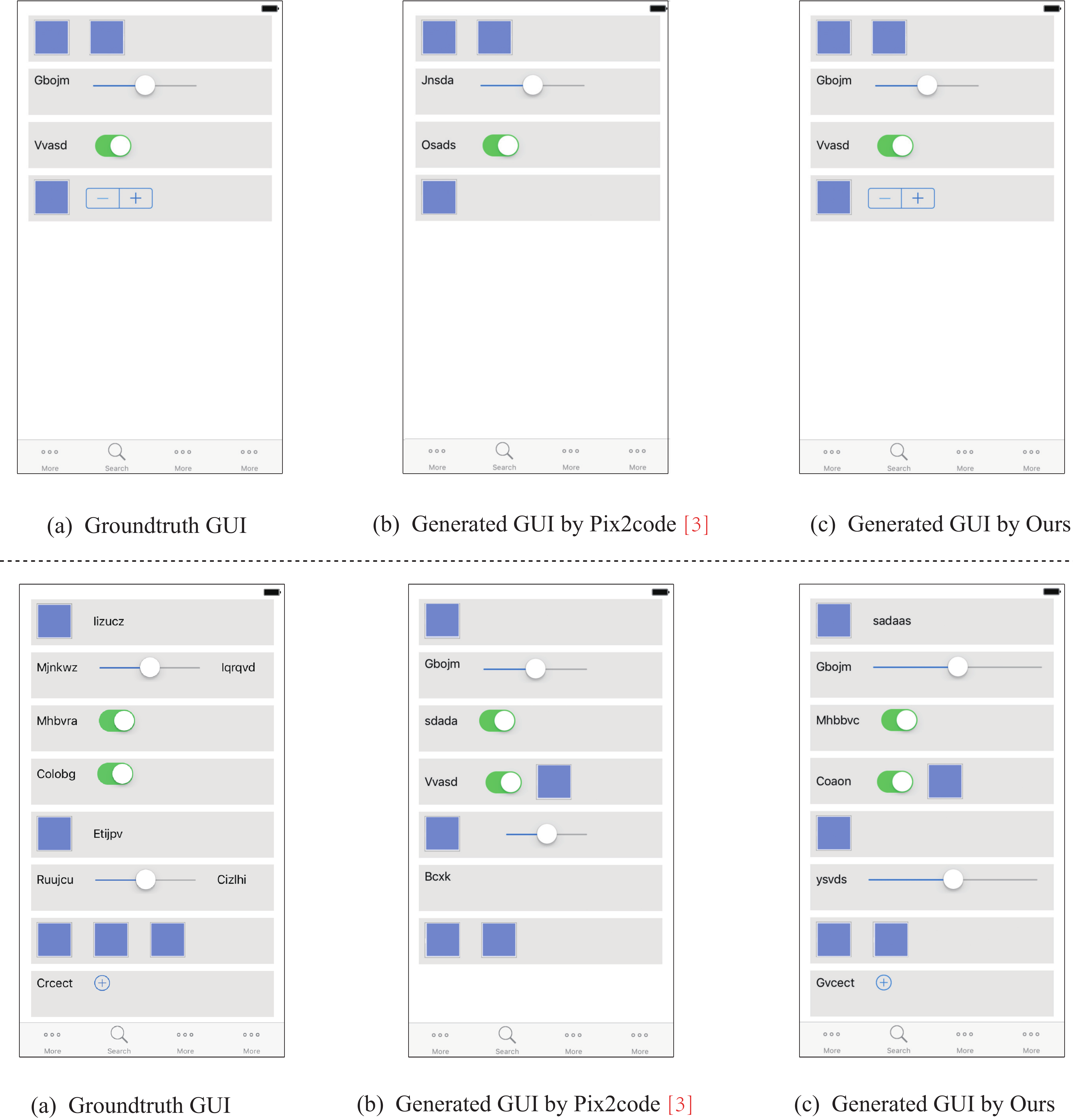}}
\caption{Experiment samples from the iOS-based GUI dataset. The examples on the upper row are form the PixCo dataset and the examples on the bottom row are from the PixCo-e dataset.}
\label{ios}
\end{figure*}

To better compare the qualitative results achieved by our method and the current state-of-the-art method, we render the DSL codes generated by two methods into GUIs. Figure \ref{web} and Figure \ref{ios} show the examples of iOS and Web examples.

Note that we use the same approach as \cite{C2} to ignore the actual textual value and assign the text contents in the label randomly. We observe that on the PixCo dataset, despite some occasional problems like misplacement and wrong styles of certain graphic elements, the results output by pix2code \cite{C2} and our model are both very close to the ground truth. This proves that two models can both learn simple GUI layout in a satisfying manner. However, on the PixCo-e dataset where the GUIs get more complicated, we note that the quality of the GUIs generated by pix2code\cite{C2} drops drastically and is clearly inferior to ours: many graphic elements from the input GUI are even not presented in the GUI generated by \cite{C2}, while our model can preserve and recover most of the visual details in the input GUI. Qualitative evaluation results on two different datasets demonstrate our model's advantage in interpreting more complex GUIs.

\section{Conclusion}

In this paper, we propose a novel method for automatically generating the code of GUI. Our method is able to achieve state-of-art performance on a benchmark GUI-code dataset as well as a dataset established by our own. Our model used a hierarchically-structured decoder integrated with attention mechanism, capable of better capturing the hierarchical layout of the GUI and the code. Our method is proven to be very effective in solving the long-term dependency problem and is able to generate codes for complex GUIs accurately. For the next steps, we plan to experiment with new methods that can recognize and segregate the overlapped graphic elements and generate the correct code to describe them.

\bibliographystyle{splncs04}
\bibliography{egbib}

\end{document}